\documentclass[conference]{IEEEtran}
\IEEEoverridecommandlockouts

\usepackage{cite}
\usepackage{amsmath,amssymb,amsfonts}
\usepackage{algorithmic}
\usepackage{graphicx}
\usepackage{textcomp}
\usepackage{xcolor}
\usepackage{booktabs}
\usepackage{url}
\usepackage{multirow}
\usepackage{subfigure}

\def\BibTeX{{\rm B\kern-.05em{\sc i\kern-.025em b}\kern-.08em
    T\kern-.1667em\lower.7ex\hbox{E}\kern-.125emX}}
\begin{document}

\title{DVM: Towards Controllable LLM Agents in Social Deduction Games}

\author{\IEEEauthorblockN{Zheng Zhang\textsuperscript{1*}\thanks{\textsuperscript{*}Equal contribution.}, Yihuai Lan\textsuperscript{1*}, Yangsen Chen\textsuperscript{1}, Lei Wang\textsuperscript{2}, Xiang Wang\textsuperscript{3}, Hao Wang\textsuperscript{1\dag}\thanks{\textsuperscript{\dag}Corresponding author.}}
\IEEEauthorblockA{\textsuperscript{1}\textit{The Hong Kong University of Science and Technology (Guangzhou)}, Guangzhou, China \\
\textsuperscript{2}\textit{Singapore Management University}, Singapore \\
\textsuperscript{3}\textit{University of Science and Technology of China}, Hefei, China \\
Email: zzhang302@connect.hkust-gz.edu.cn, haowang@hkust-gz.edu.cn}
}

\maketitle

\begin{abstract}
Large Language Models (LLMs) have advanced the capability of game agents in social deduction games (SDGs). These games rely heavily on conversation-driven interactions and require agents to infer, make decisions, and express based on such information. While this progress leads to more sophisticated and strategic non-player characters (NPCs) in SDGs, there exists a need to control the proficiency of these agents. This control not only ensures that NPCs can adapt to varying difficulty levels during gameplay, but also provides insights into the safety and fairness of LLM agents. In this paper, we present DVM, a novel framework for developing controllable LLM agents for SDGs, and demonstrate its implementation on one of the most popular SDGs, Werewolf. DVM comprises three main components: Predictor, Decider, and Discussor. By integrating reinforcement learning with a win rate-constrained decision chain reward mechanism, we enable agents to dynamically adjust their gameplay proficiency to achieve specified win rates. Experiments show that DVM not only outperforms existing methods in the Werewolf game, but also successfully modulates its performance levels to meet predefined win rate targets. These results pave the way for LLM agents' adaptive and balanced gameplay in SDGs, opening new avenues for research in controllable game agents.
\end{abstract}





\section{Introduction}
\label{sec:intro}
Large Language Models (LLMs), owing to their exceptional understanding, planning, and decision-making capabilities \cite{brown2020language, chowdhery2023palm, touvron2023llama, eigner2024determinant}, have demonstrated impressive performance in various gaming environments \cite{hu2024survey, tan2024cradle, qin2024mp5, wang2023describe, wang2023voyager, cloos2024baba}. Among them, social deduction games (SDGs), which emphasize logical reasoning, strategic thinking and conversation skills, present a critical challenge for LLM agents \cite{light2023from, akata2023playing, chi2024amongagents, kim2024microscopic}. Previous works on agents playing SDGs primarily focused on maximizing performance, striving to achieve the highest possible win rates. Xu et al. \cite{xu2023language} proposed a framework that integrates LLMs with reinforcement learning to develop strategic language agents for the Werewolf game. Wu et al. \cite{wu2024enhance} introduced a Thinker module to enhance the reasoning capabilities of LLM agents in complex deduction tasks. Lipinski et al. \cite{lipinski2022emergent} and Xu et al. \cite{xu2023exploring} explored emergent communication strategies and tuning-free frameworks for LLMs in SDGs.
Besides, Lai et al. \cite{lai-etal-2023-werewolf} and Zhang et al. \cite{werewolfxl} have investigated multimodal approaches, which combined visual signals with dialogue context to model persuasion behaviors.
These advancements underscore the growing integration of LLMs in enhancing game agents' performance.

However, this singular focus on optimal performance often overlooks the importance of having controllable agent performance during gameplay. Enabling agents to adjust their gameplay proficiency to match specified skill levels is beneficial for dynamic difficulty scaling, aligning with the actual requirements of game development. Despite these advantages, research on controllable LLM agents that can dynamically adjust their performance levels remains relatively unexplored.

This paper introduces DVM (\underline{\textbf{D}}ynamic \underline{\textbf{V}}ictory \underline{\textbf{M}}anager), a framework for LLM agents to dynamically adjust gameplay proficiency using reinforcement learning, maintaining a specified win rate for controllable performance. DVM comprises three components: Predictor, Decider, and Discussor. The Predictor aids in understanding player relationships by analyzing interactions and behaviors during gameplay. The Decider leverages the Predictor's insights and game state information to make strategic decisions. And the Discussor produces contextually appropriate and strategic dialogues to influence other players.

\begin{figure*}
  \centering
  \includegraphics[width=0.85\linewidth]{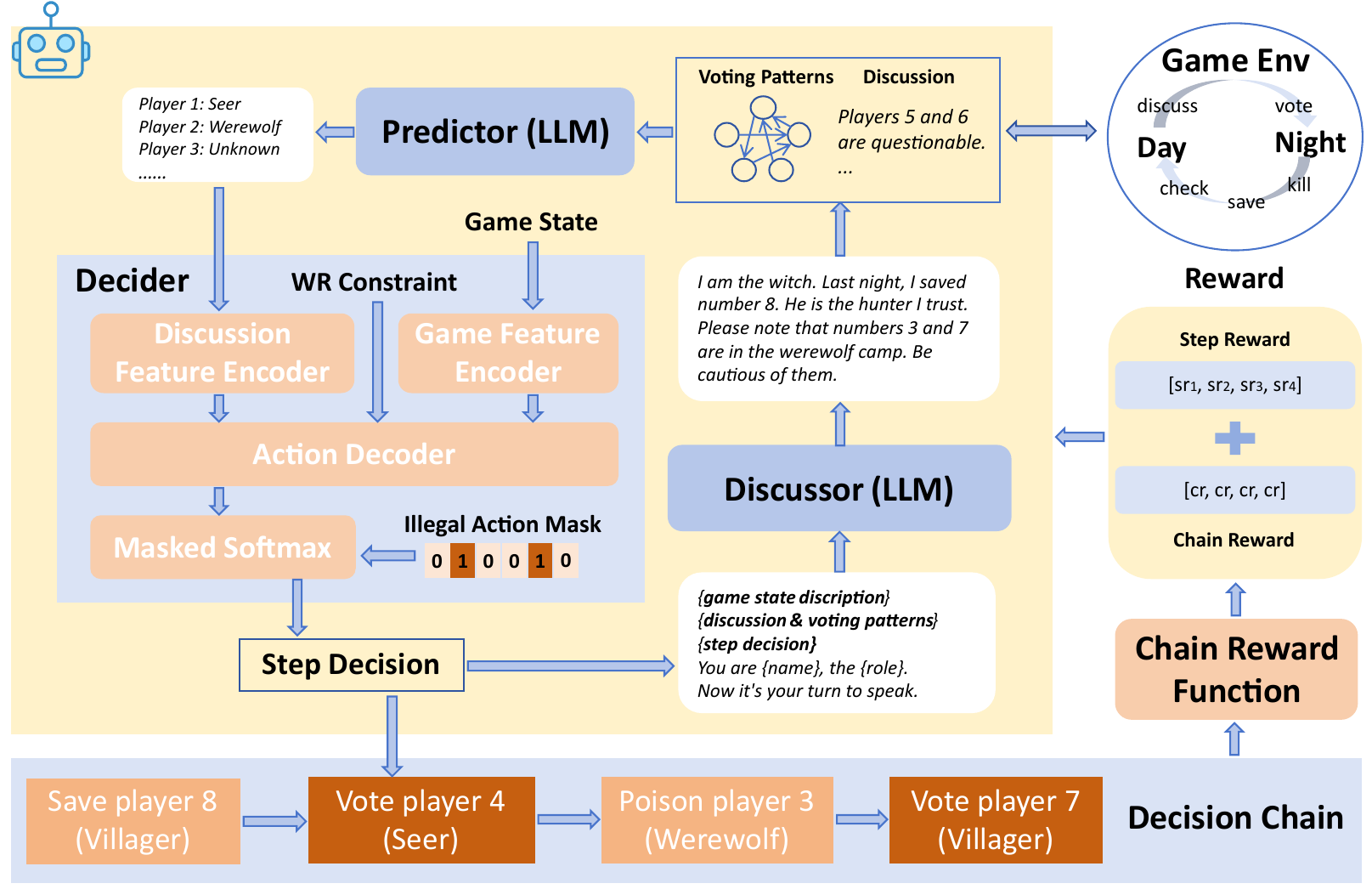}
  \caption{\textbf{The framework of DVM.} DVM consists of three parts: Predictor, Decider, and Discussor. The final reward is obtained by adding the step reward and the decision chain reward.}
  \label{fig:framewark}
\end{figure*}

The training of the Decider employs the PPO algorithm \cite{schulman2017ppo}, known for its stability in complex environments \cite{zhang2019proximal, ZHANG2022750, zheng2023secrets, xu2024dposuperiorppollm, zhong2023a}. A novel decision chain reward, which evaluates the quality of entire decision sequences rather than individual actions, is introduced to enhance strategic performance. Furthermore, DVM incorporates a win rate-constrained reward function that adjusts based on the deviation from a specified win rate, supporting the development of a controllable agent.

The principal contributions of this work are as follows:
\begin{itemize}
\setlength{\itemsep}{0pt}
\setlength{\parskip}{0pt}
\setlength{\parsep}{0pt}
\item We propose DVM, a comprehensive framework with prediction, decision and discussion modules for reinforcement learning in social deduction games.
\item We develop a win rate-constrained decision chain reward function, enabling agents to dynamically adjust the proficiency to maintain a specified win rate.
\item We conducted experiments in the Werewolf game and demonstrate that DVM not only surpasses current methods but also successfully adjusts its performance to achieve specific win rate goals.
\end{itemize}





\section{DVM}
\label{gen_inst}


This section introduces the components of DVM, followed by a detailed description of the training methods for each module. The framework is shown in Fig. \ref{fig:framewark}.

\subsection{Agent Components}
\textbf{Predictor.} In social deduction games (SDGs), understanding the intentions and roles of other players is crucial to making strategic decisions. The Predictor analyzes players' interactions and behaviors to provide insights into relationships among players, aiding in more informed decision-making and enhancing cooperation with allies and confrontation with adversaries. Initially proposed with a dense network \cite{serrino2019finding}, we introduce an LLM-based Predictor to differentiate teammates from opponents using current discussion and voting patterns. The output of the Predictor informs the decision-making process, formulated as:
\[
P_t = \text{Predictor}(D_t, V_t)
\]

where \( D_t \) represents prior discussion content, \( V_t \) represents prior voting patterns at game phase \( t \), and \( P_t \) represents predictions about the identities of other players. These predictions are expressed in formatted natural language, then parsed and vectorized for the Decider.

\textbf{Decider.} The Decider decides agent actions based on observations and the Predictor's predictions. It uses three embedding layers to encode the subject, verb, and object of each event, and can output actions like voting, checking, saving, or killing, depending on the game phase. The goal is to optimize agent survival and success, considering roles and threats. This module calculates action probabilities using the Softmax function, formulated as:
\[
\text{Logits}(a) = \text{Decider}(G_t, P_t, WR_{cons.})
\]
\[
\text{Prob}(a) = \text{Softmax}(\text{Logits}(a) - a_{mask} \times 10^9)
\]

where \(a\) is the action to be taken, \(G_t\) is the current game state, \(P_t\) is the Predictor's predictions, \(WR_{cons.}\) is the win rate constraint, and \(a_{mask}\) represents the illegal action mask based on game rules.



\textbf{Discussor.} The task of the Discussor is to produce coherent and contextually appropriate speech for the agents during the discussion phase of the game. The generated text aims to persuade, deceive, or inform other agents based on the agent's strategy and role. The Discussor receives input from the Decider's decision along with the current game context described in natural language. The output is in text format, which is then used in the game environment to simulate realistic discussions among the agents. We represent the discussion speech and voting patterns at phase \(t\) using the following equation:
\[
\text{\(D_{t+1}\)} = \text{Discussor}(G_t, D_t, V_t, S_t) + D_t
\]
\[
\text{\(V_{t+1}\)} = \text{Discussor}(G_t, D_t, V_t, S_t) + V_t
\]

where \(G_t\) is the current game state, \(D_t\) is the prior discussion content, \(V_t\) is the voting patterns, and \(S_t\) is the step decision made by the Decider.

\subsection{Training Methods}

In our framework, we utilize ChatGLM3-6B \cite{glm2024chatglm} as the base model for both the Predictor and the Discussor. However, we only train the Predictor and the Decider. The training process is executed in two steps: supervised training and reinforcement learning.

\textbf{Supervised Training.} 
Initially, we conduct supervised training on the FanLang-9 dataset\footnote{\url{https://github.com/boluoweifenda/werewolf}}, which consists of over 18,000 records of human-player games. This phase aims to fine-tune the Predictor and train the Decider to understand the intricacies of player decisions and strategies within the Werewolf environment.

\textbf{Reinforcement Learning.}
Following supervised training, we further optimize the Predictor and the Decider using reinforcement learning. Each component undergoes a training process through self-play to refine their policies. For the Predictor, we use its predictions and the correct answers as negative and positive samples to train it using DPO \cite{rafailov2023direct}. For the Decider, we employ the PPO algorithm \cite{schulman2017ppo} due to its stability and efficiency in handling complex policies. The optimization objective of PPO is formulated as:

\[
\mathcal{L}_{PPO}(\theta) = -\mathbb{E}_{s, a \sim \pi_{\theta'}} \left[ \frac{\pi_\theta(a|s)}{\pi_{\theta'}(a|s)} A^{\pi_{\theta}}(s, a) \right]
\]
\[
A^{\pi_{\theta}}(s, a) = r_t + \gamma V(s_{t+1}) - V(s_t)
\]

Where \(\pi_{\theta}\) is the current policy, \(\pi_{\theta'}\) is the previous policy, and \(r_t\) is the reward at step t.

To better capture long-term strategies, we introduce a decision chain reward mechanism to account for overall game-level decisions rather than single-step rewards. Therefore, \(r_t\) is calculated as:
\[
r_t = sr_t + cr
\]

Where \(sr_t\) is the step reward, and \(cr\) is the chain reward. The combined reward \(r_t\) is then used to optimize the Decider.

\textbf{Decision Chain Reward.}
We enhance agent training by evaluating decision chains across the entire gameplay, as opposed to focusing only on single-step rewards used by Wu et al.\cite{wu2024enhance}.

We analyzed decision chains from the FanLang-9 dataset, and calculated the win rate for each chain, creating a database of \((DC, WR)\) pairs, where \(DC\) represents the decision chain and \(WR\) represents the win rate. After each game, the win rate of the agent’s decision chain is retrieved from this database, providing either positive or negative rewards:
\[
cr(DC) = \alpha \cdot (WR - 0.5)
\]

where \( \alpha \) is a constant that controls the amplitude.

This reward promotes strategic sophistication by rewarding sequences of decisions that lead to higher win rates, thus improving the agent's game performance.

\textbf{Controllable Agent Training.}
The training of controllable agents focuses on developing game agents that can not only perform tasks effectively but also be adjusted to maintain a controllable win rate. Such a strategy ensures stability across games by not always making the best or worst decisions, aligning with the practical needs of difficulty regulation in game development.

To achieve this, we redefine the decision chain reward of the Decider with a function tied to the win rate constraint. The difference between the win rate constraint and the current win rate is calculated, and a threshold \(\epsilon\) determines the sign of the reward. The win rate difference is then converted into a controlled reward \(cr_{ctrl.}\) and scaled within \([-s, s]\) using a factor \(s\) as follows:
\[
d = \left( WR_{cons.} - WR_{dc} \right)^2
\]
\[
r = -\tanh\left( \frac{d - \epsilon^2}{k} \right)
\]
\[
cr_{ctrl.} = 
\begin{cases} 
r \cdot \left( 1 - \frac{d}{\epsilon} \right) \cdot s & \text{if } r \geq 0 \\
r \cdot \frac{d - \epsilon}{1 - \epsilon} \cdot s & \text{if } r < 0
\end{cases}
\]

where \(k\) is the smoothing factor for the tanh function.

This reward function is designed to encourage the agent to stabilize its win rate by providing positive rewards for small deviations and negative rewards for large deviations from the win rate constraint. The scaling of rewards within a specific range enhances gradient information, guiding the agent's learning process.

\section{Experiments}
\label{experiment}
This section describes the experimental details to evaluate the performance of DVM. We conducted experiments comparing the DVM with different LLM agents in playing the Werewolf game. All these games were conducted in a 9-player setup, consisting of 3 werewolves, 1 seer, 1 witch, 1 hunter, and 3 villagers. The evaluation focuses on two main aspects: the controllability of the agents and the overall performance of the framework.

\begin{figure*}
    \centering
    \subfigure[ReAct \cite{yao2022react}]{
        \includegraphics[width=0.23\linewidth]{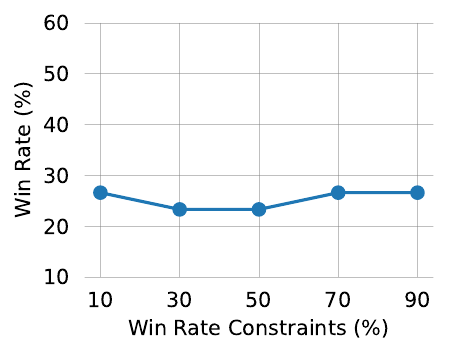}
    }
    \subfigure[LtM \cite{zhou2023leasttomost}]{
        \includegraphics[width=0.23\linewidth]{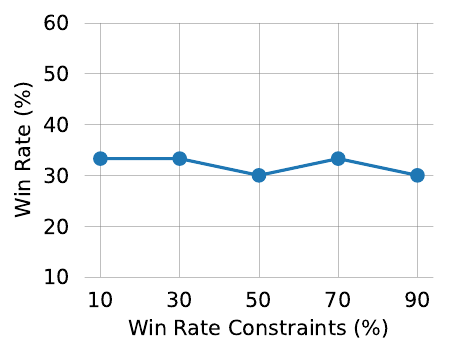}
    }
    \subfigure[Thinker \cite{wu2024enhance}]{
        \includegraphics[width=0.23\linewidth]{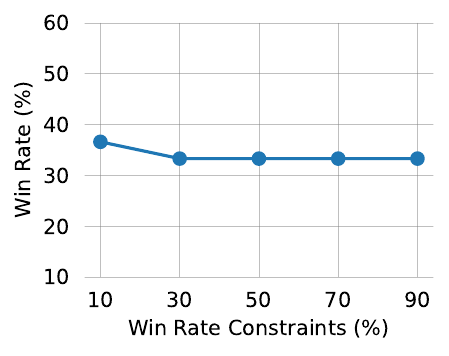}
    }
    \subfigure[DVM (ours)]{
        \includegraphics[width=0.23\linewidth]{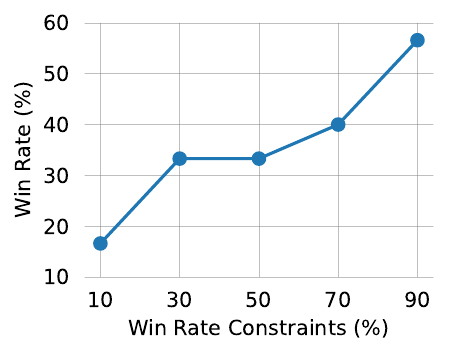}
    }
    \caption{\textbf{Controllability performance of agents.} For each method, we applied it to control the village side in the game and added different win rate constraints. The other roles was controlled by Thinker. We conducted 30 games under each setting and measured the actual win rate for the village side.}
    \label{fig:winrate}
    \vspace{-5px}
\end{figure*}

\begin{table*}
  \caption{\textbf{Prediction performance.} Werewolf Prediction means identifying 3 werewolves out of the other 8 players. Identity Prediction means predicting identities of all other 8 players. ACC@N is the probability that at least N predictions are correct.}
  \label{tab:role_prediction}
  \centering
  \setlength{\tabcolsep}{12pt}
  \begin{tabular}{lcccccc}
    \toprule
    \multirow{2}{*}{Method} & \multicolumn{3}{c}{Werewolf Prediction} & \multicolumn{3}{c}{Identity Prediction} \\
    \cmidrule(r){2-4} \cmidrule(r){5-7}
    & ACC@1 & ACC@2 & ACC@3 & ACC@1 & ACC@3 & ACC@5 \\
    \midrule
    Random & 0.748 & 0.206 & 0.008 & 0.908 & 0.344 & 0.031 \\
    GPT-3.5 & 0.725 & 0.239 & 0 & 0.951 & 0.478 & 0.065 \\
    GPT-4 & 0.805 & 0.306 & 0.041 & 0.958 & 0.592 & 0.153 \\
    LtM \cite{zhou2023leasttomost} & 0.842 & 0.321 & 0.045 & 0.955 & 0.592 & 0.158 \\
    Thinker \cite{wu2024enhance} & 0.853  & 0.326 & 0.015 & 0.963 & 0.600 & 0.158  \\
    DVM (ours) & \textbf{0.908} & \textbf{0.462} & \textbf{0.090} & \textbf{0.972} & \textbf{0.633} & \textbf{0.170}  \\
    \bottomrule
  \end{tabular}
  \vspace{-13px}
\end{table*}

\subsection{Controllability of the Agent}

The controllability of the agent is evaluated by comparing the actual win rates achieved by the agents against the specified win rate constraints. Fig. \ref{fig:winrate} illustrates the performance of different agents under different win rate constraints. In our DVM, we adjusted the win rate constraint provided to the Decider. For the other methods, we simply incorporated the constraints into their prompts in the form of text.




The result indicates that the other methods are not sensitive to the win rate constraints specified in their prompts. Their actual win rates show little fluctuation regardless of changes to these constraints. In contrast, our method demonstrates a noticeable upward trend in actual win rates as the win rate constraints increase, even though there remains a gap between the achieved win rates and the target constraints. This trend indicates that our agents exhibit a degree of controllability.

Additionally, we observed that our agents perform more effectively when targeting lower win rate constraints. However, as the win rate constraints are raised, it becomes increasingly challenging for our agent to meet these higher targets. This observation is understandable, given the intrinsic difficulty of pushing an agent to exceed its top performance limits. Our main goal is to adjust the agent's capabilities to achieve a specific win rate that is beneath its optimal potential, thereby showcasing our ability to control its performance.

\begin{table}
  \caption{\textbf{Performance Comparison.} For the evaluation of each method, we used it to control roles in a certain camp, while the other roles were controlled by Thinker.}
  \label{tab:comparison_other_methods}
  \centering
  \begin{tabular}{lccc}
    \toprule
    Method & Werewolf & Villager & Other Roles \\
    \midrule
    ReAct \cite{yao2022react} & 53.3 & 26.6 & 23.3 \\
    GPT-3.5 & 53.3 & 30.0 & 26.6 \\
    GPT-4 & 60.0 & 33.3 & 36.6 \\
    LtM \cite{zhou2023leasttomost} & 60.0 & 33.3 & 33.3 \\
    Thinker \cite{wu2024enhance} & 63.3 & 36.6 & 36.6 \\
    DVM (ours) & \textbf{66.6} & \textbf{63.3} & \textbf{53.3} \\
    \bottomrule
  \end{tabular}
  \vspace{-8px}
\end{table}

\subsection{Overall Performance of Agent Framework}

We selected 600 games from the FanLang-9 dataset and from games played by different agents. We then randomly sampled discussion content and voting patterns from these games as our test set. We conducted a comparative analysis between our DVM and other methods to assess their performance in predicting players' identities. As detailed in Table \ref{tab:role_prediction}, our framework surpasses other methods across two prediction tasks.

Besides, we selected Thinker as the baseline and evaluated the win rates of different methods against Thinker under unrestricted conditions. For DVM, we set the win rate constraint to 100\%. We conducted 30 games under each setting. The results in Table \ref{tab:comparison_other_methods} show that our proposed framework outperforms other methods, and the ablation study in Table \ref{tab:ablation_study} demonstrates the effectiveness of each component. Specifically, our method achieves a win rate of 63.3\% for the villager, 66.6\% for the werewolf, and 53.3\% for the other roles.

\section{Conclusion}

\begin{table}
  \caption{\textbf{Ablation study.} DCR represents decision chain reward.}
  \label{tab:ablation_study}
  \centering
  \begin{tabular}{lccc}
    \toprule
    Method & Werewolf & Villager & Other Roles \\
    \midrule
    DVM (ours) & \textbf{66.6} & \textbf{63.3} & \textbf{53.3} \\
    \quad -w/o. DCR & 63.6 & 56.6 & 50.0 \\
    \quad -w/o. Predictor & 63.6 & 46.0 & 40.0 \\
    \bottomrule
  \end{tabular}
  \vspace{-8px}
\end{table}


In this paper, we introduced DVM, a novel framework for creating controllable LLM agents in SDGs like Werewolf. DVM features three core components: Predictor, Decider, and Discussor, each contributing to the agent's analytical, decision-making, and communication skills. By integrating reinforcement learning with a unique decision chain reward mechanism and a win rate-constrained reward function, DVM allows agents to dynamically adjust their gameplay proficiency to meet specified win rates. Experimental results show that DVM outperforms existing methods and successfully maintains target win rates, advancing LLM agents' adaptive and balanced gameplay in SDGs and providing new insights into the development of controllable LLM agents.

\section*{Acknowledgments}

This research is supported by SMP-IDATA Open Youth Fund, the National Natural Science Foundation of China (No. 62406267), Guangzhou-HKUST(GZ) Joint Funding Program (Grant No.2023A03J0008), Education Bureau of Guangzhou Municipality and the Guangzhou Municipal Education Project (No. 2024312122).

\bibliographystyle{IEEEbib}
\bibliography{strings,refs}

\end{document}